%% file: arxiv_version.tex
\documentclass{article}
\usepackage{spconf}

\usepackage{xcolor}
\usepackage{tikz}
\usetikzlibrary{spy,arrows}
\usepackage{pgfplots}
\usepackage{bbm}
\usepackage[ruled,vlined]{algorithm2e}
\SetKwInput{kwInit}{Initialization}
\usepackage{amsmath,epsfig,amssymb,amsthm,cite,url,xcolor}
\usepackage{enumerate}

\usepackage[acronym]{glossaries}
\glsdisablehyper
\loadglsentries{glossary.tex}

\theoremstyle{plain}
\newtheorem{theorem}{Theorem}
\newtheorem{assumption}{Assumption}

\theoremstyle{definition}

\newcommand{\ex}{{\mathbb E}}

\usepackage{graphicx}
\usepackage{placeins}
\usepackage{float}
\usepackage{tabularx}
\usepackage{tkz-euclide,subfigure}
\usepackage{epsfig,amssymb}
\usepackage{bbm}
\usetikzlibrary{intersections}
\usepgfplotslibrary{fillbetween}

\usepackage{mathtools}
\pgfplotsset{compat=1.12}
\usepackage{svg}
\usepackage{hyperref}

\DeclareUnicodeCharacter{0080}{}
\UseRawInputEncoding

\title{Multi-Fidelity Hybrid Reinforcement Learning via Information Gain Maximization}
\name{Houssem Sifaou and Osvaldo Simeone\thanks{
        The work of O. Simeone was  supported by the Open Fellowships of the EPSRC (EP/W024101/1) and by the EPSRC project (EP/X011852/1).}}
\address{KCLIP, CIIPS, Department of Engineering, King's College London, London, UK}
\begin{document}
%
\maketitle
\begin{abstract}
Optimizing a reinforcement learning (RL) policy typically requires extensive interactions with a high-fidelity simulator of the environment, which are often costly or impractical. Offline RL addresses this problem by allowing training from pre-collected data, but its effectiveness is strongly constrained by the size and quality of the dataset. Hybrid offline-online RL leverages both offline data and interactions with a single simulator of the environment. In many real-world scenarios, however, multiple simulators with varying levels of fidelity and computational cost are available. In this work, we study multi-fidelity hybrid RL for policy optimization under a fixed cost budget. We introduce multi-fidelity hybrid RL via information gain maximization (MF-HRL-IGM), a hybrid offline-online RL algorithm that implements fidelity selection based on information gain maximization through a bootstrapping approach. Theoretical analysis establishes the no-regret property of MF-HRL-IGM, while empirical evaluations demonstrate its superior performance compared to existing benchmarks.
\end{abstract}

\glsresetall
\section{Introduction}
\label{sec:intro}

\begin{figure}[ht]
  \centering
  \includegraphics[width=0.4\textwidth]{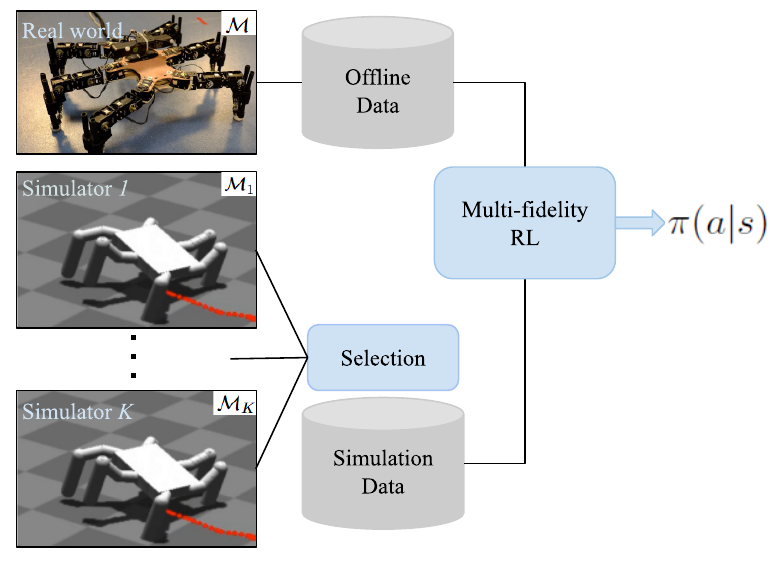}
  \caption{Multi-fidelity hybrid \gls{rl} setting: The goal is to learn an optimal policy for the real environment $\mathcal{M}$, given an offline dataset collected in $\mathcal{M}$ and access to $K$ simulators of varying fidelity and cost, which can be leveraged to generate additional simulation data.}
  \label{fig:overview}
\end{figure}


Reinforcement learning (RL) typically requires millions of interactions with a high-fidelity simulator~\cite{Kaiser2020Model}. While low-fidelity simulators are computationally cheaper, they introduce performance degradations due to the sim-to-real gap~\cite{peng2018sim,niu2022trust,ruah2023bayesian}. Offline RL addresses this challenge by training policies on pre-collected datasets, thereby eliminating the need for online simulator interactions~\cite{levine2020offline}. It has demonstrated success in applications such as interactive recommendation~\cite{xiao2021general}, control~\cite{zhan2022deepthermal,eldeeb2024conservative}, and robotics~\cite{lee2021beyond}. However, its effectiveness critically depends on the size and coverage of the offline dataset in the state–action space~\cite{levine2020offline}.

\begin{figure*}[t]
\begin{equation}
\begin{aligned}
\min_{Q}\;\; & \underbrace{\alpha_c \,\left(\log\sum_{s,a}\omega(s,a)\exp\bigl(Q(s,a)\bigr)\;-\;\mathbb{E}_{s,a\sim\mathcal{D}^{\text{off}}}\bigl[Q(s,a)\bigr]\right)}_{\text{Conservative value regularization}}\\
&\;+\;\underbrace{\mathbb{E}_{(s,a,s')\sim\mathcal{D}^{\text{off}}}\bigl[(Q - \mathcal{T}^{\pi}Q)(s,a)\bigr]^{2}
\;+\;\mathbb{E}_{(s,a,s')\sim\mathcal{B}}\left[
\frac{P_{\mathcal{M}}(s'\!\mid\!s,a)}{P_{\widehat{\mathcal{M}}}(s'\!\mid\!s,a)}
\bigl[(Q - {\mathcal{T}}^\pi Q)(s,a)\bigr]^{2}\right]}_{\text{Bellman error on offline and online data}}.
\end{aligned}
\label{eq:h2o_objective}
\end{equation}
\end{figure*}

A promising direction is to combine offline RL with online RL in imperfect simulators, exploiting the strengths of both paradigms --  reducing reliance on large offline datasets, while avoiding costly interactions with high-quality simulators~\cite{niu2022trust,niu2023h2o+,hou2024improving}. This motivation has driven the development of hybrid offline-online RL methods. Existing work, however, largely assumes access to a single-fidelity simulator, neglecting the trade-off between simulator fidelity and computational cost. As shown in conventional RL, multi-fidelity simulators can provide varying levels of accuracy at different costs, enabling more efficient use of computational resources~\cite{cutler2014reinforcement,suryan2020multifidelity,khairy2024multi,liu2025multifidelity}.

In this work, we introduce multi-fidelity hybrid RL, focusing on the adaptive selection of simulator fidelity levels under a fixed cost budget. Our method, multi-fidelity hybrid RL via information gain maximization (MF-HRL-IGM), is inspired by bootstrap RL~\cite{hu2024bayesian} and multi-fidelity Bayesian optimization~\cite{song2019general}, employing the information gain per unit cost as a principled criterion for fidelity selection. Our main contributions are:
\begin{itemize}
\item We formulate, for the first time, the problem of fidelity-level selection in hybrid RL, integrating bootstrapped RL for uncertainty quantification~\cite{hu2024bayesian} with multi-fidelity selection strategies based on information gain~\cite{song2019general}.
\item We provide a theoretical regret analysis of the proposed algorithm.
\item We validate our approach experimentally on a standard RL environment, showing superior performance over existing benchmarks.
\end{itemize}


\section{Problem Definition}
\label{sec:pr_def}
We consider a classical \gls{mdp} setting defined by the tuple $
\mathcal{M} = \bigl(\mathcal{S}, \mathcal{A}, r, P_{\mathcal{M}}, \rho, \gamma\bigr),
$ where $\mathcal S$ denotes the state space, $\mathcal{A}$ is the action space, $r(s,a) \in [0,1]$ is the reward function, $P_{\mathcal{M}}(s' | s,a)$ is transition probability, $\rho(s)$ is the initial state distribution, and $\gamma \in (0,1]$ is a discount factor. We address the conventional objective of finding a policy $\pi(a| s)$ that maximizes the expected discounted return
\begin{equation}
J_{\mathcal{M}}(\pi) =
\mathbb{E}_{\substack{s_0 \sim \rho,\, a_t \sim \pi(\cdot|s_t),\\ s_{t+1}\sim P_{\mathcal{M}}(\cdot|s_t,a_t)}}
\Bigg[\sum_{t=0}^H \gamma^t\,r(s_t,a_t)\Bigg],
\label{eq:obj}
\end{equation}
where $H$ denotes the finite horizon of the \gls{mdp}. 

As illustrated in Fig.~\ref{fig:overview}, we study policy optimization when direct interaction with the true environment $\mathcal{M}$ is unavailable, but we have access to $K$ \emph{simulators} $\{\mathcal{M}_k\}_{k=1}^K$ of the true environment. Each simulator $\mathcal{M}_k$ is characterized by a tuple $\bigl(\mathcal{S}, \mathcal{A}, r_k, P_{\mathcal{M}_k}, \rho_k, \gamma\bigr)$, which generally differ from the true environment $\mathcal{M}$. In particular, each simulator $\mathcal{M}_k$ is characterized by a \emph{fidelity} $l_k$ and \emph{cost} $\lambda_k$ satisfying the inequalities $l_1 < l_2 < \cdots < l_K$ and $\lambda_1 > \lambda_2 > \cdots > \lambda_K$. Accordingly, the simulators $\{\mathcal{M}_k\}_{k=1}^K$ provide increasingly accurate approximations of the true environment $\mathcal{M}$ as index $k$ increases, but the increased accuracy comes at a larger cost. In addition to the possibility to use the $K$ simulators, we are also given a fixed offline dataset $\mathcal{D}^{\text{off}} = \{(s_i,a_i,r_i,s'_i)\}_{i=1}^n$, which is collected under the true dynamics $\mathcal{M}$ using a fixed unknown behavioral policy $\pi_\beta$. 

Our goal is to learn a policy $\pi$ that maximizes the expected discounted return \eqref{eq:obj} under the true environment $\mathcal{M}$, while imposing a constraint on simulation budget. Specifically, for any run of the policy optimization process, we impose the inequality 
$
\sum_{t=1}^T \lambda_{k_t}\leq \Gamma$ over  the optimization horizon $T\gg H$, where  $k_t\in \{1,\cdots,K\}$ is the chosen simulator index at time $t$, and $\Gamma$ is the overall cost budget.
\section{Background}
\label{sec:background}
In this section, we briefly review online and offline \gls{rl} and introduce hybrid offline-online \gls{rl}, which serves as a starting point for the solution to the multi-fidelity hybrid RL problem addressed in this paper.
\vspace{-0.2cm}
\subsection{Online and Offline RL}
Standard actor-critic RL algorithms alternate between policy evaluation and policy improvement (see, e.g.,~\cite{Sutton1998}). We distinguish between online and offline \gls{rl} algorithms. In \emph{online RL}, the data $\{(s,a,r,s')\}$ used for policy evaluation and policy improvement is collected from the true environment $\mathcal{M}$ using previous iterates of the policy $\pi$. In contrast, with \emph{offline RL}, one can only rely on a fixed offline dataset $\mathcal{D}^{\text{off}} = \{(s_i,a_i,r_i,s'_i)\}_{i=1}^n$ generated by following a behavioral policy~\cite{levine2020offline}. 

To elaborate on this, define the Bellman evaluation operator under policy $\pi$ as
\begin{align}
\mathcal{T}^{\pi}\widehat Q(s,a)
= r(s,a) \;+\;\gamma\,\mathbb{E}_{a'\sim\pi(\cdot\mid s')}\bigl[\widehat Q(s',a')\bigr],
\label{eq:bellman}
\end{align}
where $\widehat Q(s,a)$ is an estimate of the action-value function $Q(s,a) = \ex [\sum_{t=0}^H \gamma^{t} r(s_t,a_t)]$ for policy $\pi$, where the average is taken as in~\eqref{eq:obj}. The action-value function estimate $\widehat Q(s,a)$ is optimized by minimizing the expected Bellman residual as
\begin{align}
 &\widehat Q \;\leftarrow\; \arg\min_{Q}\;\mathbb{E}_{s,a,s'\sim\mathcal U}\Bigl[\bigl(Q(s,a) - \mathcal{T}^{\hat\pi}\widehat Q(s,a)\bigr)^2\Bigr],
 \label{eq:policy_eval}
 \end{align}
and the policy is updated by acting greedily with respect to function $\widehat Q(s,a)$ as
\begin{align}
\hat\pi \;\leftarrow\; \arg\max_{\pi}\;\mathbb{E}_{s,a\sim\mathcal U}\bigl[\widehat Q(s,a)\bigr].
 \label{eq:policy_imp}
\end{align}
In \eqref{eq:policy_eval}-\eqref{eq:policy_imp}, the notation $\mathcal U$ represents either a replay buffer $\mathcal B$ collected using previous iterates of the policy for online RL, or the fixed offline dataset $\mathcal{D}^{\text{off}} = \{(s_i,a_i,r_i, s_{i}')\}_{i=1}^n$ for offline RL. For the latter, the objectives in \eqref{eq:policy_eval} and \eqref{eq:policy_imp} typically include additional regularization terms to mitigate the risk of overestimating the action-value function based on offline data~\cite{levine2020offline}. This is further discussed next.
\vspace{-0.2cm}
\subsection{Hybrid Offline-Online RL}
Real-world decision-making often requires leveraging both offline data and online interactions to achieve sample-efficient and optimal policy learning. To this end, reference~\cite{niu2022trust} proposed the H2O algorithm, which leverages online data generated by a single simulator ${\mathcal{M}}_1$. 


H2O builds upon \emph{conservative Q-learning} (CQL)~\cite{kumar2020conservative}, optimizing the objective in \eqref{eq:h2o_objective}. In \eqref{eq:h2o_objective}, the parameter $\alpha_c>0$ controls the strength of the regularization term that penalizes overestimation of the action-value function. This is done by contrasting high-density Q-values against the empirical mean over the offline dataset $\mathcal{D}^{\text{off}}$. The importance weight $\omega(s,a)$ in \eqref {eq:h2o_objective} is derived from the normalized KL divergence between real and simulated transition distributions, namely $P_{\mathcal{M}}(\cdot\!\mid\!s,a)$ and $P_{{\mathcal{M}}_1}(\cdot\!\mid\!s,a)$, i.e., $    \omega(s,a)\,\propto\,D_{\mathrm{KL}}(P_{\mathcal{M}}(\cdot\!\mid\!s,a)\,\|\,P_{{\mathcal{M}}_1}(\cdot\!\mid\!s,a))
$, and is normalized over all visited state-action pairs.

 The second term in \eqref{eq:h2o_objective} combines the conventional squared Bellman error on offline transitions $\mathcal{D}^{\text{off}}$ with an importance-weighted Bellman error in \eqref{eq:policy_eval} on simulated data $\mathcal{B}$. The dynamics ratio $P_{\mathcal{M}}(s'|s,a)/P_{\widehat{\mathcal{M}}}(s'|s,a)$, which corrects for the simulator bias, can be rewritten as \cite[Eq. 7]{niu2022trust}
  \begin{align}
      \tfrac{P_{\mathcal{M}}(s'|s,a)}{P_{\widehat{\mathcal{M}}}(s'|s,a)}
      = \frac{P(\text{real}|s,a,s')(1- P(\text{real}|s,a))}{P(\text{real}|s,a)(1- P(\text{real}|s,a, s'))},
  \end{align}
   where $P(\text{real}|s,a,s')$ and $P(\text{real}|s,a)$ denote the posterior probability that the real dynamics $\mathcal{M}$ has generated the given observations $(s,a,s')$ and $(s,a)$, respectively. 
  In practice, the probabilities $P(\text{real}|s,a)$ and $P(\text{real}|s,a,s')$ are approximated using learned discriminators that are optimized using cross-entropy loss between real and simulated data~\cite{eysenbach2021offdynamics}.

\vspace{-0.3cm}
\section{Multi-fidelity Hybrid RL}
\label{sec:proposed_method}

In this section, we introduce MF-HRL-IGM, a hybrid offline-online \gls{rl} algorithm that extends H2O to incorporate a mechanism for the selection of the simulator fidelity level.\\
\textbf{Overview:} MF-HRL-IGM trains $L$ hybrid-\gls{rl} policies simultaneously, allowing for uncertainty quantification~\cite{hu2024bayesian} and dynamic fidelity level selection. The approach consists of two phases: 1) an \emph{offline phase} during which $L$ bootstrapped datasets are generated from the offline dataset $\mathcal{D}^{\text{off}}$, supporting the optimization of $L$ policies; and 2) an \emph{online phase} during which the $L$ hybrid-\gls{rl} policies are trained using the offline dataset along with new transitions generated via the simulators $\{\mathcal{M}_k\}_{k=1}^K$ at different fidelity levels. The fidelity level is chosen dynamically by maximizing the information gain per unit cost~\cite {song2019general}. \\
{\textbf{Offline Phase:}} As in \cite{hu2024bayesian}, $L$ resampling masks $\{M_{\ell}\}_{\ell=1}^{L}$ are generated to create bootstrapped datasets from the offline dataset $\mathcal{D}^{\text{off}}$. Each binary mask $M_{\ell} \in \{0,1\}^n$ determines which entries from the original dataset $\mathcal{D}^{\text{off}}$ are included in bootstrapped subset $\mathcal{D}_{\ell}^{\text{off}}$. These masks are sampled independently, with entries given by i.i.d. Bernoulli random variables with parameter $0.5$. The bootstrapped datasets are used to train $L$ offline policies and action-value functions $\{\pi_\ell^{\text{off}}, Q_\ell^{\text{off}}\}_{\ell=1}^L$ using the CQL algorithm \cite{kumar2020conservative}.\\
{\textbf{Online Phase:}} During this phase, we start from the pretrained $L$ offline policies, and refine them based on new interactions generated in the simulators
, without access to the true environment.  
We proceed in \emph{rounds} $r\in\{1,\cdots, R\}$, where each round $r$ consists of collecting new trajectories in one of the $K$ simulators, denoted as $k_r\in\{1,\cdots,K\}$, and performing policy updates using H2O.

To effectively select the fidelity level at which the simulator is run, we maintain a \emph{generalized posterior belief}~\cite{bissiri2016general,simeone2022machine,knoblauch2022optimization,zecchin2023robust} over the set of candidate policies $\{\pi_{\phi_\ell}\}_{\ell=1}^L$, at each round $r$. Let $U \in \{1,\dots,L\}$ denote a latent random variable indexing the best policy within the candidate set. 
Our goal is to maintain and update the posterior distribution $p(U \mid \mathcal{D}^r)$ as new data becomes available, where $\mathcal{D}^r$ is the data available during round $r$, including the offline dataset $\mathcal{D}^{\text{off}}$. We set $\mathcal{D}^0 = \mathcal{D}^{\text{off}}$.

To elaborate, define as $\mathcal{L}_\ell( \mathcal{B}_k)$ the H2O loss in~\eqref{eq:h2o_objective} corresponding to policy $\ell$ when evaluated with the $k$-th simulator data $\mathcal{B}_k$. Then, given the available data $\mathcal{D}^{r-1}$ at the beginning of round $r$, including the offline dataset and the freshly collected data $\mathcal{B}_{k_r}^r$ at the current round, the posterior over the $L$ policies is updated as
    \begin{align}
        p(U=\ell|\mathcal{D}^r) \propto  p(U=\ell|\mathcal{D}^{r-1}) \exp(-\mathcal{L}_\ell( \mathcal{B}_{k_r}^r)).
        \label{eq:posterior}
    \end{align}
 At the start of each round $r$, we decide the fidelity level based on the information gain per unit cost. To this end, we define as $\mathbb{I}(U; \mathcal{\mathcal B}_k^r|\mathcal{D}^{r-1})$ the mutual information between latent variable $U$ and simulated data $\mathcal{\mathcal B}_k^r$ conditioned on data $\mathcal{D}^{r-1}$. This term quantifies the information obtained about the optimal policy index $U$ by adding data ${\mathcal B}_k^r$ collected at fidelity level $k$.

The fidelity level at round $r$ is then selected to maximize the information gain per unit cost
    \begin{align}
        k_r  = \max_{k} \frac{\mathbb{I}(U; \mathcal{\mathcal B}_k^r|\mathcal{D}^{r-1})}{\lambda_k},
        \label{eq:criterion}
    \end{align}
as long as the condition
\begin{align}
\frac{\mathbb{I}(U;\mathcal{B}_{k_r}^r|\mathcal{D}^{r-1}) }{\lambda_{k_r}}\geq \beta_r,
\label{eq:cond}
\end{align}
is satisfied for a threshold $\beta_r$. If condition \eqref{eq:cond} is not satisfied, the highest fidelity level is selected, i.e., $k_r = K$. Following the regret analysis in Sec~\ref{sec:regret_section}, we specifically set $\beta_r = 1/\sqrt{\Gamma_r}$, where $\Gamma_r$ is the remaining cost budget at round $r$, i.e., $\Gamma_r = \Gamma -\sum_{i=1}^{r-1}\lambda_i$. 

To estimate the mutual information in \eqref{eq:criterion}, we express it as the difference between the entropy of the posterior, $\mathbb{H}(U | \mathcal{D}^{r-1})$, and the expected conditional entropy, \\$\mathbb{E}_{\mathcal{B}_k^r}[\mathbb{H}(U |\mathcal{B}_k^r, \mathcal{D}^{r-1})]$~\cite{cover1999elements}, which are estimated as in~\cite{paninski2003estimation}. To this end, for the first term, the posterior is updated with the newly collected data $\mathcal{B}_{k_r}^r$, while for the second term, the posterior is approximated using historical simulation batches $\mathcal{B}_k$ for each fidelity level $k$.
\vspace{-0.3cm}
\section{Regret Analysis}
\label{sec:regret_section}
In this section, we study the regret of MF-HRL-IGM. First, following~\cite{xu2021fine,jin2020provably}, we define a notion of regret tailored to our multi-fidelity RL setting. Specifically, we define the multi-fidelity regret as
\begin{equation}
\mathcal{R}(\Gamma)\!=\!N_e\!\left[\tfrac{\Gamma}{\lambda_K}\,
\ex_{s_0}\!\left[V^{\pi^\star}(s_0)\right]\!-\!
\sum_{r=1}^R \ex_{s_0}\!\left[V_{k_r}^{\pi_r}(s_0)\right]\right]
\label{eq:regret}
\end{equation}
where $N_e$ represents the fixed number of episodes run at each round $r$; $V^{\pi^\star}(s_0)$ is the expected return in \eqref{eq:obj} under the optimal policy $\pi^\star$; and $V_{k_r}^{\pi_r}(s_0)$ is the expected return under current policy $\pi_r$.  The first term in \eqref{eq:regret} represents the optimal return when the maximum number of episodes are run at the highest fidelity level $K$ under cost-budget $\Gamma$, while the second term amounts to the sum of all returns under the current policy and fidelity level selection strategy. A multi-fidelity optimization algorithm is called \emph{no-regret} if $\lim_{\Gamma \to \infty} {\mathcal{R}(\Gamma)}/{\Gamma} = 0$ \cite{song2019general}.
Henceforth, we write $g^\star = N_e \ex_{s_0}\left[V^{\pi^\star}(s_0)\right]$ and $g_{k_r}^r = N_e \ex_{s_0}\left[V^{\pi_r}_{k_r}(s_0)\right]$, for ease of notation.
\begin{assumption}
We assume the following: (\emph{i}) The highest fidelity level $K$ corresponds to the true environment, i.e., $\mathcal{M}_K = \mathcal{M}$.  (\emph{ii}) Additional data collected under lower fidelity levels and additional policy updates do not worsen the regret, i.e.,
    $
        \sum_{r=1}^R\mathbbm{1}\{k_r = K\}(g^\star- g_{K}^r)\leq \sum_{r=1}^{\tilde R} (g^\star - \tilde{g}^r)
        $,    
    with $\tilde R =  \sum_{r=1}^R\mathbbm{1}\{k_r = K\}$ and $\tilde{g}^r$ is the average return corresponding to a policy trained by using only fidelity level $K$ for $\tilde R$ rounds.
    (\emph{iii}) The linear MDP condition in \cite[Assumption A]{jin2020provably} holds, and \cite[Algorithm 1]{jin2020provably} is employed for policy optimization.
  \label{assump1}  
\end{assumption}
The following theorem shows that MF-HRL-IGM is a no-regret algorithm. The proof can be found in the Appendix.
\begin{theorem}
     Under Assumption \ref{assump1}, the regret in \eqref{eq:regret} satisfies the upper bound
\begin{align}
    \mathcal{R}(\Gamma)  \leq C\alpha(\Gamma) \gamma_{\rm low} + \mathcal{O}(\sqrt{ \Gamma} \log(\Gamma)),
    \label{main_result}
\end{align} 
where $\gamma_{\rm low} = \sum_{r=1, k_r\neq K}^R {\mathbb{I}(U;\mathcal{B}_{k_r}^r|\mathcal{D}^{r-1}) }$, $C=g^*/\lambda_K$, and $\alpha(\Gamma) = \max_{r}1/\beta_r$.
 In particular, with threshold $\beta_r=1/\sqrt{\Gamma_r}$ in \eqref{eq:cond}, the regret grows sublinearly with respect to the cost budget  $\Gamma$ as  
$    \mathcal{R}(\Gamma)   = \mathcal{O}(\sqrt{ \Gamma} \log(\Gamma))$. 
\label{thm1}
\end{theorem}
\vspace{-0.3cm}
\section{Numerical Results}
\label{sec:num_results}
In this section, we present an empirical evaluation of the proposed MF-HRL-IGM framework using the standard MuJoCo control suite. Specifically, we consider the Half-Cheetah environment as the true environment, while lower-fidelity simulators are constructed by perturbing key dynamics parameters, such as gravity and friction~\cite{niu2022trust,niu2023h2o+}. Specifically, we construct three lower-fidelity simulators by scaling the original gravity by factors of $[2.75, 2.0, 1.25]$. For the offline dataset, we utilize $250\text{K}$ samples from the \emph{halfcheetah-medium-replay-v0} dataset~\cite{kumar2020conservative}. We consider $L=3$ bootstrap policies, which are trained using CQL \cite{kumar2020conservative} for $100$ epochs during the offline phase and further optimized with H2O ~\cite{niu2022trust} for $500$ epochs during the online phase.

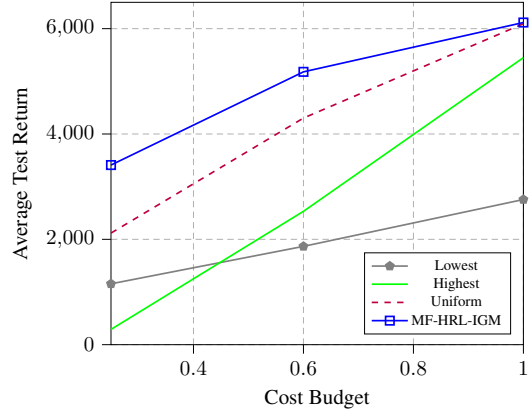
\begin{figure}[t]
\begin{center}
\begin{tikzpicture}[scale=0.8]
\begin{axis}[
tick align=outside,
tick pos=left,
x grid style={white!69.0196078431373!black},
xlabel={Cost Budget},
xmajorgrids,
xmin= 0.25, xmax=1.00,
xtick style={color=black},
y grid style={white!69.0196078431373!black},
ylabel={Average Test Return},
ymajorgrids,
ymin=0.0, ymax=6500.0,
ytick style={color=black},
grid=major,
scaled ticks = true,
legend pos = south east,
legend style={nodes={scale=0.7, transform shape}},
grid style=densely dashed,
]
\addplot  [thick, color = gray, mark = pentagon*, mark size = 2, mark repeat = 0, mark phase = 0]
coordinates {
(0.25, 1155.5)(0.6, 1866.3)(1.0, 2756.0)
}; \addlegendentry{Lowest}
\addplot  [thick, color = green, mark = none, mark size = 2, mark repeat = 0, mark phase = 0]
coordinates {(0.25, 289.40)(0.6, 2531.6)(1.0, 5449.5)
}; \addlegendentry{Highest}
\addplot  [thick, color = purple, dashed , mark =none,  mark size = 2, mark repeat = 0, mark phase = 0]
coordinates {(0.25, 2118.6)(0.6, 4302.31)(1.0, 6093.4)
}; \addlegendentry{Uniform}
\addplot  [thick, color = blue, mark = square, mark size = 2, mark repeat = 0, mark phase = 0]
coordinates {
(0.25, 3410.34)(0.6, 5179.9)(1.0, 6116.5)
}; \addlegendentry{MF-HRL-IGM}
\end{axis}
\end{tikzpicture}
\end{center}
\vspace{-0.5cm}
\caption{Average test return as a function of the cost budget for the proposed MF-HRL-IGM strategy against benchmarks based on fixed fidelity levels or on a uniform cost allocation across fidelity levels with increasing fidelity across rounds.}
\label{fig:gravity}
\end{figure}

In Fig.~\ref{fig:gravity}, we report the average test return of the proposed MF-HRL-IGM method against three benchmark strategies: (\emph{i}) always selecting the lowest-fidelity level; (\emph{ii}) always selecting the highest-fidelity level, and (\emph{iii}) uniformly distributing the cost budget across all fidelity levels, so that the fidelity level increases across the round index $r$. The results in the figure show that MF-HRL-IGM consistently achieves the highest performance across different budget settings. When the budget is large, uniform allocation performs competitively, approaching the optimal return. However, under more restrictive cost budgets, its performance degrades significantly compared to MF-HRL-IGM. Furthermore, committing to a single fidelity level proves to be clearly suboptimal, highlighting the benefit of adaptively balancing across different fidelity levels.

\section{Conclusion}
\label{sec:ccl}
In this paper, we introduced a multi-fidelity hybrid \gls{rl} framework that leverages information gain to handle settings with only offline data and simulators of varying fidelity and cost. We established a no-regret guarantee for the proposed selection scheme and validated its effectiveness through theoretical analysis and experiments. While our empirical evaluation focused on the H2O algorithm, the approach is general and can be applied to other hybrid \gls{rl} methods. Extending the framework to additional algorithms and more complex environments is a promising direction for future research.

{\small
\bibliographystyle{IEEEbib}
\bibliography{strings,references}}

 \onecolumn
\section*{Appendix}
\label{appendix}
 With the notations $g^\star = N_e \ex_{s_0}\left[V^{\pi^\star}(s_0)\right]$ and $g_{k_r}^r = N_e \ex_{s_0}\left[V^{\pi_r}_{k_r}(s_0)\right]$, the regret can be written as
\begin{align}
\mathcal{R}(\Gamma) &= \frac{\Gamma}{\lambda_K} g^* -\sum_{r=1}^R g_{k_r}^r\nonumber\\
    & \overset{(a)}{\leq}\frac{\Gamma}{\lambda_K} g^*  - \sum_{r=1}^R \mathbbm{1}\{k_r=K\} g_{K}^r\nonumber\\
    &\leq \left(\frac{\Gamma}{\lambda_K} - \sum_{r=1}^R \mathbbm{1}\{k_r=K\}\right)g^* +\sum_{r=1}^R \mathbbm{1}\{k_r=K\}\left( g^* - g_K^r\right)\nonumber\\ 
     &\overset{(b)}{\leq} \frac{g^*}{\lambda_K}\sum_{r=1}^R \lambda_{k_r}\mathbbm{1}\{k_r\neq K\}  +\sum_{r=1}^R \mathbbm{1}\{k_r=K\}\left( g^* - g_K^r\right)  
     \label{derivation}
\end{align} 
where $(a)$ follows from the fact that $g_{k_r}^r\geq 0$ (non-negative rewards assumption) and $(b)$ is obtained by noting that ${\Gamma} - \sum_{r=1}^R \lambda_K \mathbbm{1}\{k_r=K\} =\sum_{r=1}^R \lambda_{k_r}\mathbbm{1}\{k_r\neq K\}$. Recall that at each round $r$, we select $k_r$ following \eqref{eq:criterion} and if $k_r \neq K$, we further verify if it satisfies condition \eqref{eq:cond}.
Using \eqref{eq:cond}, the first term in \eqref{derivation} can be bounded as
\begin{align}
 \frac{g^*}{\lambda_K}\sum_{r=1}^R \lambda_{k_r}\mathbbm{1}\{k_r\neq K\}  \leq  \frac{g^*}{\lambda_K}\sum_{r=1, k_r\neq K}^R \frac{\mathbb{I}(U;\mathcal{B}_{k_r}^r|\mathcal{D}^{r-1}) }{\beta_r}\leq \frac{g^*}{\lambda_K}\alpha(\Gamma) \gamma_{\rm low}
 \label{first_term_bound}
\end{align} 
where $\alpha(\Gamma) = \max_{r}1/\beta_r$ and $\gamma_{\rm low} = \sum_{r=1, k_r\neq K}^R {\mathbb{I}(U;\mathcal{B}_{k_r}^r|\mathcal{D}^{r-1}) }$.

Now, we bound the second term in \eqref{derivation} in accordance with the single-fidelity \gls{rl} literature. From conditions (i) and (ii) in Assumption \ref{assump1}, it follows that
\begin{align}
\sum_{r=1 }^R\mathbbm{1}\{k_r=K\} \left( g^* - g_K^r\right) \leq \sum_{r=1}^{\tilde R} (g^\star - \tilde{g}^r),
\label{second_term_first_res}
\end{align}
where $\tilde{g}^r$ is the expected return at round $r$ of a policy optimized using only the highest fidelity level simulator $K$ and $\tilde{R} = \sum_{r=1 }^R\mathbbm{1}\{k_r=K\}$. From \cite[Theorem 3.1]{jin2020provably}, for a given $\delta \in (0,1)$ and under condition $(iii)$ in Assumption \ref{assump1}, we have with probability $1-\delta$
\begin{align}
\sum_{r=1}^{\tilde R} (g^\star - \tilde{g}^r) = \mathcal{O}(\sqrt{d^3 H^3 T \xi^2}),
\end{align}
where $d$ is the feature space dimension, $T= \tilde{R} N_e H$ is the total number of timesteps, and $\xi = \log(2dT/\delta) $.  Assuming that $N_e$, $H$, $d$, and $\delta$ are fixed constants, then 
\begin{align}
\sum_{r=1}^{\tilde R} (g^\star - \tilde{g}^r) = \mathcal{O}\left(\sqrt{ T (\log(T))^2}\right) \overset{(a)}{\leq} \mathcal{O}\left(\sqrt{ \Gamma (\log(\Gamma))^2}\right)
\label{second_term_bound_prem}
\end{align}
where $(a)$ is obtained by noting that $T = \tilde{R}N_eH$ and $\tilde{R} = \frac{\left(\Gamma - \sum_{r=1}^R \lambda_{k_r}\mathbbm{1}\{k_r\neq K\}\right)}{\lambda_K}\leq \frac{\Gamma}{\lambda_K} $.
\eqref{second_term_first_res} and \eqref{second_term_bound_prem} imply
\begin{align}
\sum_{r=1 }^R\mathbbm{1}\{k_r=K\} \left( g^* - g_K^r\right) \leq \mathcal{O}\left(\sqrt{ \Gamma (\log(\Gamma))^2}\right).
\label{second_term_bound}
\end{align}
Combining \eqref{derivation}, \eqref{first_term_bound}, and \eqref{second_term_bound}, we get the desired result in Theorem \ref{thm1}.

\end{document}

%% file: glossary.tex
\newacronym{sota}{SOTA}{state-of-the-art}

\newacronym{pac}{PAC}{probably approximately correct}
\newacronym{rv}{RV}{random variable}
\newacronym{iid}{i.i.d.}{independent and identically distributed}

\newacronym{gp}{GP}{Gaussian process}
\newacronym{ai}{AI}{artificial intelligence}
\newacronym{gd}{GD}{gradient descent}
\newacronym{erm}{ERM}{empirical risk minimization}
\newacronym{mlp}{MLP}{multi-layer perceptron}
\newacronym{vae}{VAE}{variational auto-encoder}
\newacronym{cdl}{CDL}{contrastive density learning}

\newacronym{p-erm}{P-ERM}{pseudo-empirical risk minimization}

\newacronym{rl}{RL}{reinforcement learning}
\newacronym{mdp}{MDP}{Markov decision process}
\newacronym{pg}{PG}{policy gradient}
\newacronym{cql}{CQL}{conservative Q-learning}

\newacronym{bo}{BO}{Bayesian optimization}
\newacronym{mfbo}{MFBO}{multi-fidelity Bayesian optimization}

\newacronym{mab}{MAB}{multi-armed bandit}
\newacronym{cmab}{CMAB}{contextual multi-armed bandit}
\newacronym{dm}{DM}{direct method}
\newacronym{ips}{IPS}{inverse propensity score}

\newacronym{3d}{3D}{three dimensional}
\newacronym{2d}{2D}{two dimensional}
\newacronym{rt}{RT}{ray tracer}

\newacronym{tx}{Tx}{transmitter}
\newacronym{rx}{Rx}{receiver}
\newacronym{bs}{BS}{base station}
\newacronym{los}{LoS}{line of sight}
\newacronym{nlos}{NLoS}{non-line of sight}
\newacronym{mimo}{MIMO}{massive input massive output}
\newacronym{cfr}{CFR}{channel frequency response}
\newacronym{toa}{ToA}{time of arrival}
\newacronym[longplural=angles of arrival]{aoa}{AoA}{angle of arrival}
\newacronym[longplural=angles of departure]{aod}{AoD}{angle of departure}
\newacronym{xpd}{XPD}{cross-polarization discriminant}
\newacronym{ckm}{CKM}{channel knowledge map}

\newacronym{cp}{CP}{conformal prediction}

\newacronym{ppi}{PPI}{prediction-powered inference}

\newacronym{dr}{DR}{doubly-robust}
\newacronym{tdr}{TDR}{tuned doubly-robust}
\newacronym{ctdr}{CTDR}{context-tuned doubly-robust}
\newacronym{cdr}{CDR}{context-aware doubly-robust}